\documentclass[letterpaper, 10 pt, conference]{ieeeconf}  

\IEEEoverridecommandlockouts                              

\usepackage{hyperref}       
\usepackage{url}            
\usepackage{booktabs}       
\usepackage{amsfonts}       
\usepackage{nicefrac}       
\usepackage{microtype}      
\usepackage{lipsum}
\usepackage{graphics} 
\usepackage{times} 
\usepackage{amsmath} 
\usepackage{amssymb}  
\usepackage{bm}  
\usepackage{algpseudocode,algorithm,algorithmicx}
\usepackage[shortcuts,acronym]{glossaries}
\usepackage{gensymb} 
\usepackage{caption}
\usepackage{graphicx}
\usepackage{multirow}
\usepackage{array}
\usepackage{subcaption}
\usepackage{xcolor}
\usepackage[outdir=./]{epstopdf}
\newcolumntype{C}[1]{>{\centering}m{#1}}

\title{End-to-end Reinforcement Learning of Robotic Manipulation with Robust Keypoints Representation}

\author{Tianying Wang$^{1}$, En Yen Puang$^{2,1}$, Marcus Lee$^3$, Yan Wu$^{2,1}$, Wei Jing$^{4,*}$
\thanks{$^{1}$ A*STAR Institute of High Performance Computing (IHPC), Singapore.}%
\thanks{$^{2}$ A*STAR Institute of Information Research (I$^2$R), Singapore.}%
\thanks{$^{3}$ National University of Singapore (NUS), Singapore.}%
\thanks{$^{4}$ Alibaba DAMO Academy, Hangzhou, China}%
\thanks{$^{*}$ Corresponding author, email: {\tt\small 21wjing@gmail.com}}%
}

\newacronym{rl}{RL}{Reinforcement Learning}

\begin{document}
\maketitle

\begin{abstract}
We present an end-to-end \gls{rl} framework for robotic manipulation tasks, using a robust and efficient keypoints representation. The proposed method learns keypoints from camera images as the state representation, through a self-supervised autoencoder architecture. The keypoints encode the geometric information, as well as the relationship of the tool and target in a compact representation to ensure efficient and robust learning. After keypoints learning, the \gls{rl} step then learns the robot motion from the extracted keypoints state representation. The keypoints and \gls{rl} learning processes are entirely done in the simulated environment.
We demonstrate the effectiveness of the proposed method on robotic manipulation tasks including grasping and pushing, in different scenarios. We also investigate the generalization capability of the trained model. In addition to the robust keypoints representation, we further apply domain randomization and adversarial training examples to achieve zero-shot sim-to-real transfer in real-world robotic manipulation tasks. 
\end{abstract}

\section{Introduction} \label{sec:intro}

With the recent advancement in deep neural network and computational capacities, Reinforcement Learning (RL) has demonstrated the capabilities of achieving superior performance in many applications such as atari games~\cite{mnih2015human}, go~\cite{silver2017mastering}, and complex games~\cite{vinyals2019grandmaster}. \gls{rl} has been used for various robotics tasks~\cite{kober2013reinforcement}. Among the existing applications, using \gls{rl} for the robot to learn primitive manipulation tasks is particularly attractive, since it could be a promising paradigm to help the robot learn to minimize the effort of robot programming, especially in ad-hoc and high-mix-low-volume settings. Additionally, \gls{rl} is capable to handle uncertainties and novel scenarios because of its sequential decision and generalization capabilities \cite{kober2013reinforcement, mandlekar2020learning, wang2019efficient}.

End-to-end \gls{rl} learns the robot motion command with direct sensor input (typically vision) for robot manipulation task~\cite{zhu2018reinforcement, quillen2018deep, zhang2015towards}. The end-to-end approaches minimize the effort of manually handcrafting features, such that the same framework could be reused for different manipulation tasks with little modification. With the rapid advancement of deep learning in recent years, the Convolutional Neural Network (CNN) is able to extract features from the image input directly. Several end-to-end robotic \gls{rl} methods have been proposed with the deep CNNs to process the visual input and generate motor policy \cite{zhang2015towards, lee2020making}.

\begin{figure}[!thbp]
\centering
    \includegraphics[width=0.45\textwidth]{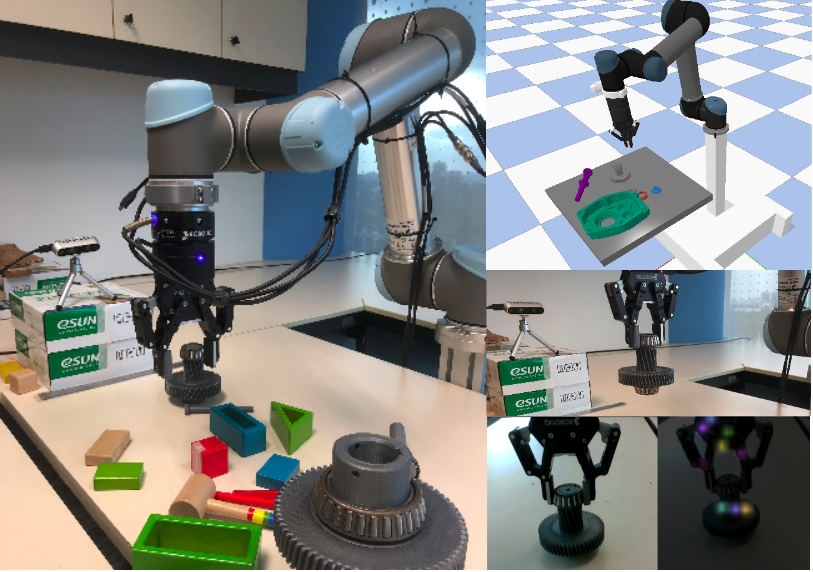}
    \caption{End-to-end RL of robotic manipulation with keypoints representation in simulation and real-world}
    \label{figure:frontimage}
    \vspace{-5.0mm}
\end{figure}

Though end-to-end robotic \gls{rl} has been studied for robotic manipulation, there are still several issues that limit its performance in real-world applications. Firstly, efficient sensory data mapping into simulation is required. Since real-world robot data collection is expensive and ad-hoc for different task settings, collecting data of direct visual input in the simulation is often desired. Furthermore, training a robust \gls{rl} control policy with direct camera input is challenging and slow to converge, especially when only simulation data is available for training. To tackle this problem, robust state representations and reliable sim-to-real transfer methods are needed.

In this work, we propose a robust keypoints representation-based RL framework for self-supervised, end-to-end robotic manipulation tasks. We use the keypoints to encode the geometric information of the input image, as well as the spatial relationship of the tool and the target in a robust and compact manner. We demonstrate that the proposed keypoints representation is able to outperform traditional methods for visual feature extraction in robotic manipulation applications. In addition, the overall RL framework is able to generalize the tasks well and achieve good success rate and fast convergence in training as well as zero-shot sim-to-real transfer in several manipulation tasks. The main contributions of this work are as follows:
\begin{itemize}
    \item We propose a novel end-to-end \gls{rl} methods with self-supervised keypoints as the state representation in robotic applications. 
    \item We achieve reliable zero-shot sim-to-real transfer of end-to-end \gls{rl} by directly applying the learnt model to real-world robot manipulation tasks.
    \item We conduct extensive experimental studies with different task scenarios in both simulation and real-world experiment, as well as studies on task generalization and transfer learning, to validate the effectiveness of the proposed algorithm. 
\end{itemize}
\begin{center}
\begin{figure*}[thpb]
     \centering
        \includegraphics[width=1\textwidth]{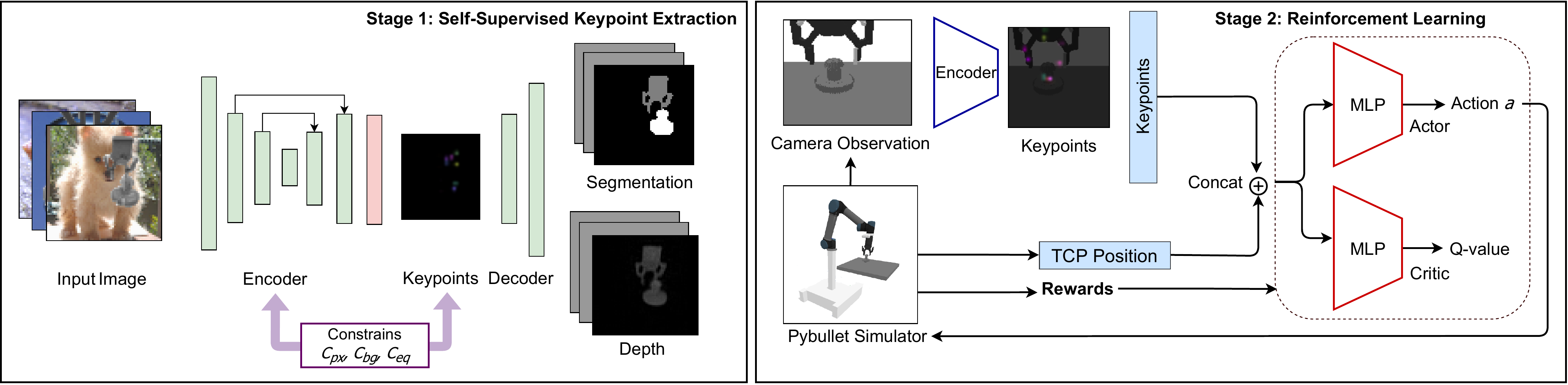}
        \caption{(Right) Overall model architecture of the proposed reinforcement learning framework with keypoints representation. (Left) Model architecture for keypoints extraction: the encoder is a DenseNet~\cite{huang2017densely} backbone with a U-Net~\cite{ronneberger2015u} architecture to enhance multi-scale feature extraction; the reconstruction objectives are depth image and semantic segmentation, with the ground truth from the simulator. We also include additional loss to constrain the formation of keypoints.}
        \label{figure:sim_training}
        \vspace{-5.0mm}
\end{figure*}
\end{center}

\section{Related Work} \label{sec:literature}

\subsection{Reinforcement Learning for Robotic Manipulation}
A number of \gls{rl} algorithms have been developed and applied to robotic manipulation in recent years~\cite{kober2013reinforcement, polydoros2017survey}. Many of these techniques are demonstrated on robot manipulation tasks using camera inputs as the sensory feedback. However, state representation and explainability  still remain as open questions \cite{heuillet2021explainability} as end-to-end \gls{rl} algorithms using images as input is difficult to train. Compressing the images to a lower dimensional meaningful state space could be helpful for faster and reliable training. \cite{raffin2019decoupling} learns feature extractors prior to \gls{rl} training to achieve good results. End-to-end task learning also requires outputting low level motion commands or trajectory from sensory input \cite{levine2016end}. Various formulations have been studied, such as end-to-end visual servoing \cite{bateux2018training, puang2020kovis} and end-to-end motor policies \cite{levine2016end, lee2020making, mahler2019learning}. \gls{rl} methods with direct visual input have also been explored \cite{zeng2018learning, joshi2020robotic, hundt2020good} for different robotic manipulation tasks. In this work, we focus on improving the end-to-end \gls{rl} with a robust keypoints representation, while the proposed method could work with most actor-critic style \gls{rl} approaches.

\subsection{Representation Learning for Visual Inputs}
Processing the direct image input is a crucial step towards end-to-end \gls{rl} in robotic manipulation. A natural approach is to use CNN to extract the visual features for direct feeding to an end-to-end \gls{rl} framework \cite{joshi2020robotic, quillen2018deep}. Using estimated poses of the objects as state for vision based \gls{rl} is also a common approach \cite{wang2019densefusion, cheng20196d}. However, it requires a good object detector and a pose estimator, which also separate the pipeline into two stages. One other approach learns a low dimensional latent space representation of the input image with an autoencoder~\cite{byravan2018se3, puang2019visual, mousavian20196}. However, these latent space features usually encode the entire physical world in the camera view which limits its generalizability. Recently, keypoint-based approaches have been used in many computer vision applications such as image generation \cite{jakab2018unsupervised}, face recognition \cite{berretti20113d}, tracking \cite{hare2012efficient} and pose estimation \cite{tulsiani2015viewpoints}. Keypoints have also been used as a representation for robotic manipulation tasks. For example, \cite{manuelli2019kpam, gao2019kpam} use keypoints for manipulation planning of category tasks with supervised learning; and \cite{qin2019keto} proposes to learn a task-specific keypoints representation for manipulation tasks via self-supervised learning. To improve the representation robustness, we propose a self-supervised keypoints representation to encode geometric information and relationship of manipulator and manipulation target. The proposed keypoints representation is an improved version of \cite{puang2020kovis}, with improved architecture and transformation loss to address the \gls{rl} framework and static camera output.

\section{Methodology}
\label{sec:method}

The overall architecture of the proposed method is shown in Fig. \ref{figure:sim_training}. The proposed method includes a keypoints extraction module using a CNN-based autoencoder with additional loss functions to capture the keypoints. The \gls{rl} module is a model-free, actor-critic style method~\cite{sutton2018reinforcement} that maps the keypoints and robot states to the motion command and Q-value. The two-stage training is conducted entirely in simulation by using the simulation ground truth to self-supervise the training. The first stage trains an autoencoder for keypoints detection, while the second stage trains the \gls{rl} module for control policy. The trained policy could be directly used in real-world experiment.

\subsection{Learning Keypoints Representation}
We use keypoints as a compact latent representation to describe the visual information of the tasks as depicted in Fig. \ref{figure:sim_training}, and hence to improve the robustness and efficiency of vision-based end-to-end \gls{rl} for the manipulation tasks. In summary, \cite{puang2020kovis} is adopted to simplify the state representation from input image to keypoints for objects in the scene of robotic manipulation task. An autoencoder is used to train the extraction of keypoints with reconstruction loss and keypoint bottle-neck. We use the ground truth of semantic segmentation and depth map as the targets for reconstruction loss. Additional losses as soft constraints are also introduced to push the keypoints away from each other, and pull them within the segmentation areas of the targets.

First, the encoder $f: \boldsymbol{x} \mapsto \boldsymbol{z}$ compresses the input image $\boldsymbol{x} \in \mathbb{R}^{H \times W \times C}$ to extract features $\boldsymbol{z} \in \mathbb{R}^{H' \times W' \times K}$. The keypoints function $\Phi: \boldsymbol{z} \mapsto \boldsymbol{k}$ then transforms the feature map into keypoints $\boldsymbol{k} \in \mathbb{R}^{K \times 3}$ by computing channel-wise centroid on the softmax of $\boldsymbol{z}$ for each of the $K$ channels:
\begin{align} \label{eqn:keypoint}
\boldsymbol{j}^*_i &= \sum_{\boldsymbol{j} \in \Omega} \boldsymbol{j} \frac{ \exp\left(\boldsymbol{z}_i\right)}{\sum_\Omega \exp\left(\boldsymbol{z}_i\right)} \\
\alpha_i &= \sigma\left(\max_\Omega\left(\boldsymbol{z}_i\right)\right)
\end{align}
where $\boldsymbol{j} = (j_1$, $j_2)$ represents the indices in the 2D axes in $\boldsymbol{z}$, and $\boldsymbol{j}^*$ is the centroid of the 2D feature map $\Omega$. Moreover, keypoint confidence $\alpha$ is computed from the sigmoid $\sigma(\cdot)$ of the channel's maximum activation. The complete keypoint representation comprises $\boldsymbol{k}_i = (\boldsymbol{j}^*_{i},\, \alpha_i)$ given $i \in [1, \dots, K]$. 

Two soft constraints are used in keypoint formation during training to improve object localization. The first constraint $\mathcal{C}_{px}$ encourages distributed keypoints formation by promoting L2-norm among centroids. The second constraint $\mathcal{C}_{bg}$ contains the keypoints within object of interest by penalizing centroid located on the background mask. 

In order to further improve the robustness and performane of keypoints learning for \gls{rl} tasks with static camera, we propose an additional transformation loss $\mathcal{C}_{eq}$ to improve consistency of keypoint extraction:
\begin{align} \label{eqn:constraint}
\mathcal{C}_{eq} = \big\| (\Phi * f * T)(\boldsymbol{x}) - (\Phi * T * f)(\boldsymbol{x}) \big\|_2
\end{align}
where $T$ is a geometrical transformation function that can be applied to both input image $\boldsymbol{x}$ and extracted feature map $\boldsymbol{z}$. This constraint requires the encoder to be consistent in object localization over geometrical transformations. It hence penalizes the differences in extracted keypoints between when transformation is applied before and after the encoder. 

\subsection{RL framework for Robotic Manipulation}

The robotic manipulation tasks discussed in this paper are formulated as Markov Decision Processes (MDPs) $M = <S, A, T, R, \gamma> $, where $S$ is a set of states, $A_i$ is a set of actions ; $T_s$ is the state transition; $R: S \times A \longrightarrow R$: is the rewards; $\gamma \in [0,1]$ is the discount factor. The total episodic return is then the summation of discounted rewards: $r_{total} = \sum_{i=0}^{T_e} \gamma^i r_i$, where $T_e$ is the total steps of an episode.

In our manipulation tasks, we use the keypoints pixel positions and robot end-effector state to represent $S$ in MDP; the action $a \in A$ is the robotic end-effector velocity command in continuous space. 
We adopt an off-policy, actor-critic style \gls{rl} algorithm \cite{sutton2018reinforcement} for the policy learning. As shown in Fig. \ref{figure:sim_training}, two parallel multilayer perceptron networks (MLPs) are used to estimate the control policy and Q-value independently. 

Our framework is suitable for any off-policy, actor-critic style RL methods, since it requires estimation of Q-value and policy, as well as efficient sampling with replay buffer. For Deep Deterministic Policy Gradient (DDPG) algorithm, the expected total return is:
\begin{equation}
    J_{\pi}(\theta) = \sum_{i=0}^{T_e} \mathbb{E}_{a,s}(\gamma^i r(a,s)) )
\end{equation}
The Q-value update is as follows:
\begin{equation}
 y_t = r(s_t, a_t) + \gamma Q_{\phi}(s_{t+1}, \pi_{\theta}(s_{t+1}))  
\end{equation}
And the policy gradient for DDPG is approximated as: 
\begin{equation}
    \nabla_{\theta} J \approx \frac{1}{N} \sum_i \nabla_{a} Q_{\phi}(s_i, a)|_{a=\pi(s_i)} \nabla_{\theta} \pi_{\theta}(s_i)
\end{equation}
where $s_t, a_t$ are the state and action at time $t$; $Q$ is the network that estimates Q-value; $\pi(s_{t+1})$ is the policy network to estimate action at state $s_{t+1}$; $y_t$ is the target to be used to train Q-network; $\theta, \phi$ are the parameters of policy network and Q network.

The Soft Actor Critic (SAC) \cite{haarnoja2018soft} includes an additional entropy item to encourage exploration:
\begin{equation}
    J_{\pi}(\theta) = \sum_{i=0}^{T_e} \mathbb{E}_{a,s}(\gamma^ir(a,s)+\alpha H(\pi(\centerdot | s)) )
\end{equation}
The target Q value is calculated as:
\begin{equation}
    y_t = r + \gamma (Q(s_{t+1}, a_{t+1}) - \alpha log \pi (s_{t+1}))
\end{equation}
Policy update could be done by taking gradient from 
\begin{equation}
    \nabla_{\theta} J \approx \nabla_{\theta}\mathbb{E}(Q_{\phi}(s, a(s,\xi))-\alpha log \pi_{\theta}(a(s,\xi|s))
\end{equation}
where $\xi$ is a fixed distribution such as multi-variant Gaussian for the stochastic policy. For simplicity, detailed formulations of DDPG and SAC are omitted.

In the training process, we use a guided reward $r_g$ to address the sparse rewards problem in the \gls{rl} training. The total reward is then denoted as $r = \alpha_1 r_g + \alpha_2 r_s$. The guided reward $r_g$ guides the target object towards the goal position, so it gives less penalty if the target object is closer to the goal position. $r_s$ is the regulation rewards which constrains the robotic end-effector to maneuver within the effective workspace. The total reward for robotic grasping task is:
\begin{equation}
\begin{split}
     r (x, y, z, d_{og}) =& -\alpha_1 r_g +\alpha_2 r_s \\
     =& -\alpha_1 max(0, min(d_{og}, 1)) + \\
      & \alpha_2(\psi(x) + \psi(y) + \psi(z))
\end{split}
\end{equation}
where $\psi(x)$ is given as:
\begin{equation}
\begin{split}
    \psi(x) = \left\{ \begin{array}{rcl}
            1 & \mbox{for} \quad x_{min} < x < x_{max} \\ 
            0 & \mbox{otherwise}
\end{array}\right.
\end{split}
\end{equation}
where $d_{og}$ is the distance between the target object position and the goal position, we clamp the guided rewards $r_g$ to better model the loss. $x$, $y$ and $z$ are the coordinates of the robotic end-effector. We use $x_{max}$, $y_{max}, z_{max}$ and $x_{min}$, $y_{min}, z_{min}$ as upper and lower bounds to softly constrain the end-effector in the effective working space.

\subsection{Network and Training Details}

\begin{figure}[!htbp]
\centering
    \includegraphics[width=0.4\textwidth]{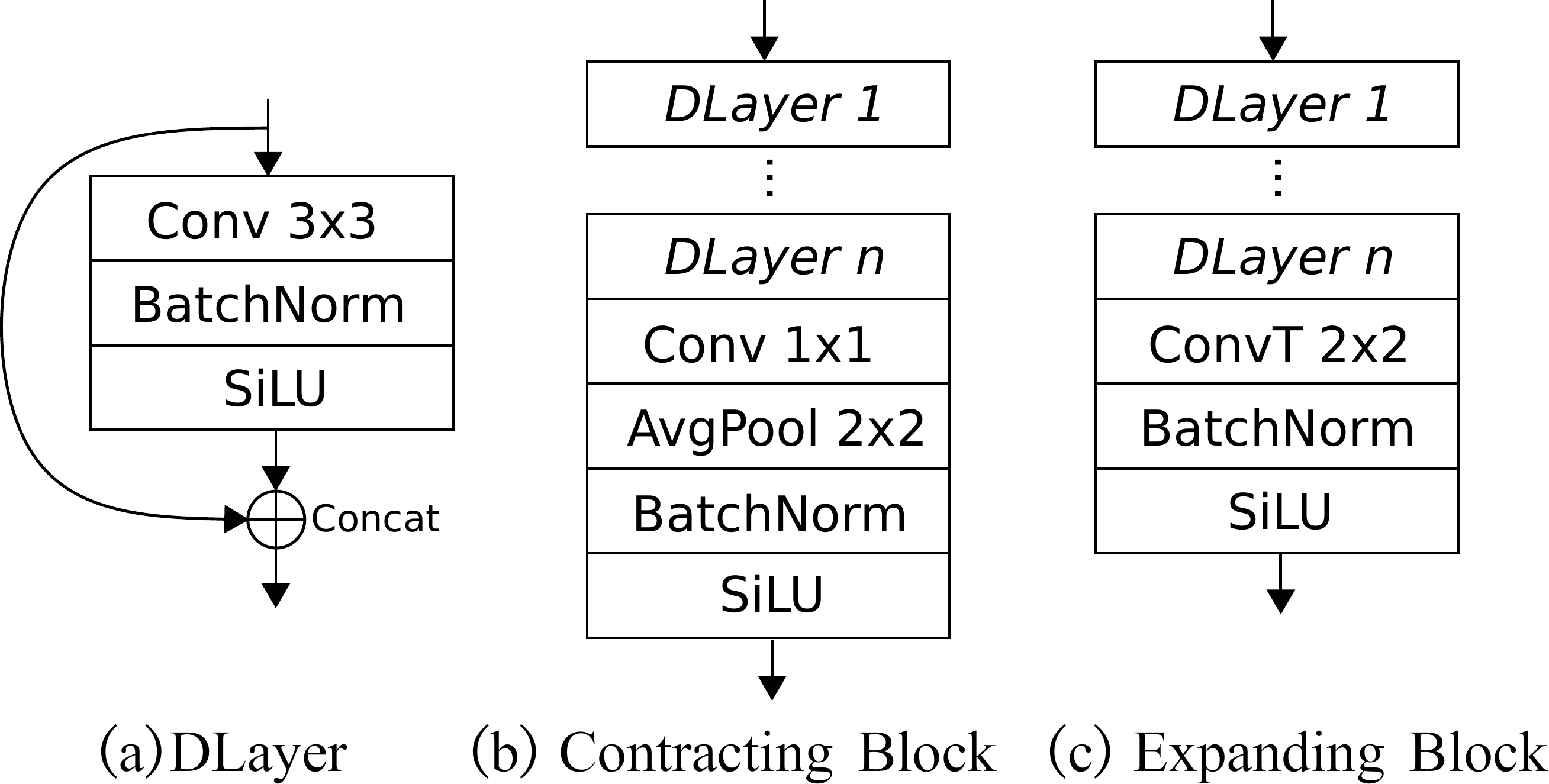}
    \caption{Configuration of each block in the DenseNet-based autoencoder for keypoints extraction. Each block contains 3 or 4 layers, the convolutional dilation increases accordingly in each layer.}
    \label{fig:densenet}
    \vspace{-5.0mm}
\end{figure}

\subsubsection{Training and Implementation of Keypoints}
For the keypoints extraction module, we adopt the autoencoder network architecture from \cite{puang2020kovis}. The keypoints extraction module consists of a fully-convolutional U-Net \cite{ronneberger2015u} with DenseNet\cite{huang2017densely} block and skip-connections in the encoder. Fig \ref{fig:densenet} depicts the configuration of blocks used in the autoencoder architecture of the keypoints extraction module. The total learnable parameters for keypoint extraction network is about 1M. 

The size of the input image $H \times W \times C$ and feature map $H' \times W' \times K$ is $64 \times 64 \times 3$ and $32 \times 32$ respectively. $K$ is the number of keypoints, which is a heuristic parameter depending on the complexity of the task. The keypoints will be used with an image pixel coordinate representation ($K \times 2$ dimension) in the later \gls{rl} stage. The geometrical transformations $T$ in (\ref{eqn:constraint}) is in-plane rotation at $90^\circ$, $180^\circ$ or $270^\circ$. We randomize the camera positions and angle within a small range, to make the algorithm more robust and calibration-free. For each task, we pre-train the keypoints extraction module, using $50,000$ images generated from the simulation. We also use domain randomization~\cite{tobin2017domain} and adversarial examples~\cite{xie2019adversarial} in the keypoints training to improve the performance and robustness for sim-to-real transfer. For the hyperparameters used during the training, the batch size is 64; the number of epochs is 16; Adam optimizer is used with learning rate of 0.001.

\subsubsection{Training and Implementation of RL}
For the reinforcement learning module, we adopt Soft Actor Critic (SAC) \cite{haarnoja2018soft} with Hindsight Experience Replay (HER) \cite{andrychowicz2017hindsight} to learn the control policy. The policy and critic networks are both $64 \times 64$ MLP networks with layer normalization. For the hyperparameters used during the training, the batch size is 512; the learning rate is 0.0003; the random exploration rate is 0.2. As for HER, the number of artificial transitions to generate for each actual transition is 4, and the buffer size is 50,000. All training experiments are conducted using several dual Nvidia GTX 2080Ti GPU workstations with Intel i9-9900X CPU. 

\section{Experiments and Results} \label{sec:res}

\begin{figure}[!thbp]
\centering
    \includegraphics[width=0.4\textwidth]{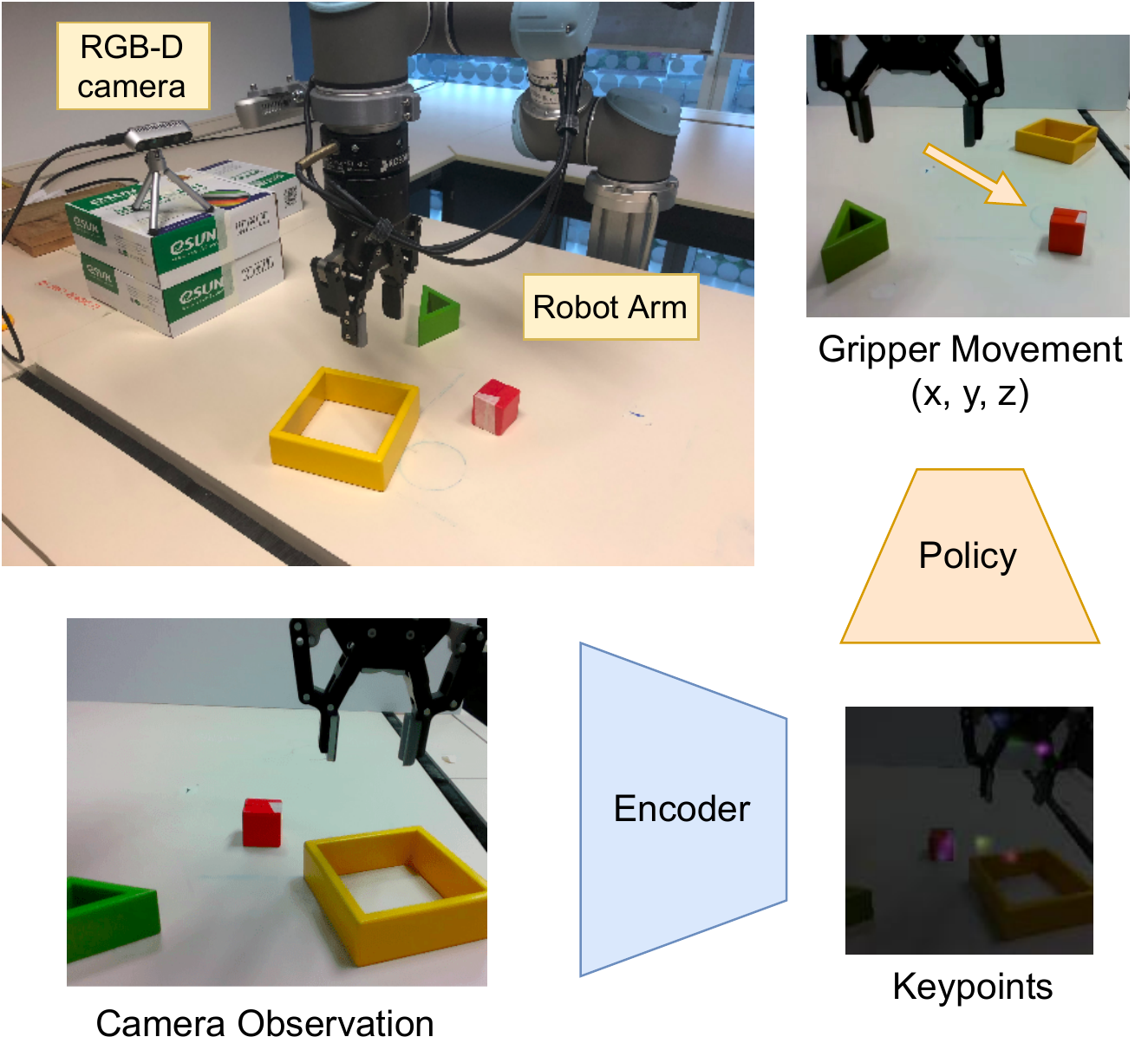}
    \caption{Experiment setup and inference pipeline}
    \label{figure:exp}
    \vspace{-4.0mm}
\end{figure}

\begin{figure}[!thbp]
\centering
    \includegraphics[width=0.4\textwidth]{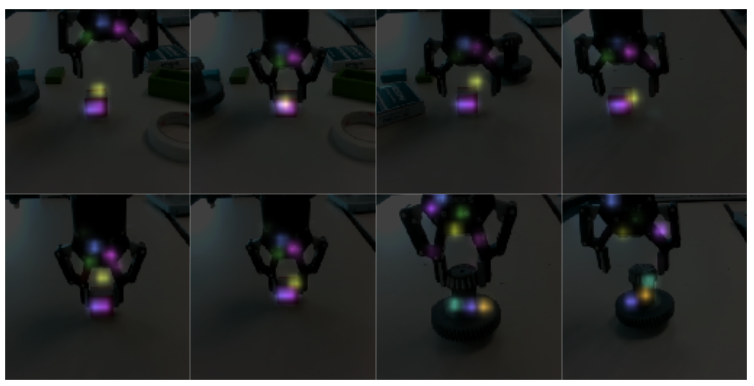}
    \caption{Keypoints extracted from the camera observations in real-world experiment}
    \label{figure:kp_sample}
    \vspace{-4.0mm}
\end{figure}

\begin{figure}[!thbp]
\centering
    \includegraphics[width=0.4\textwidth]{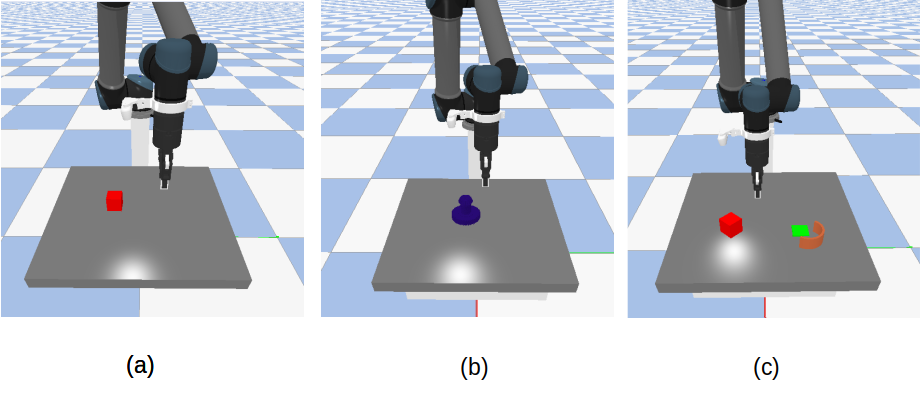}
    \caption{Setup of simulation environment: (a) \textit{UR5Grasp} env-1 (b) \textit{UR5Grasp} env-2 (c) \textit{UR5Push} env-1}
    \label{figure:tasks}
    \vspace{-5.0mm}
\end{figure}

\begin{figure*}[!thbp]
\centering
  \begin{subfigure}[b]{0.28\linewidth}
    \includegraphics[width=\textwidth]{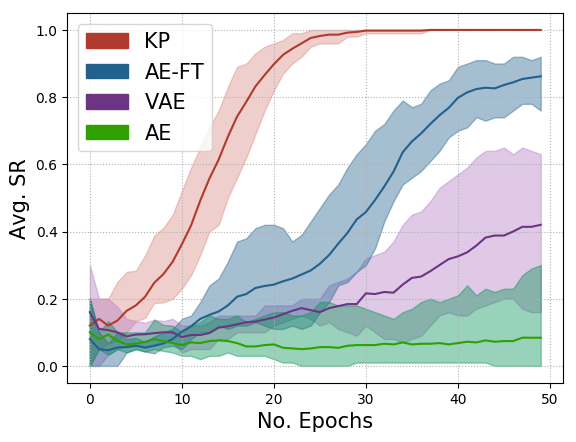}
    \caption{Success rates of \textit{UR5Grasp} env-1}
    \label{f1}
  \end{subfigure}
  \begin{subfigure}[b]{0.28\linewidth}
    \includegraphics[width=\textwidth]{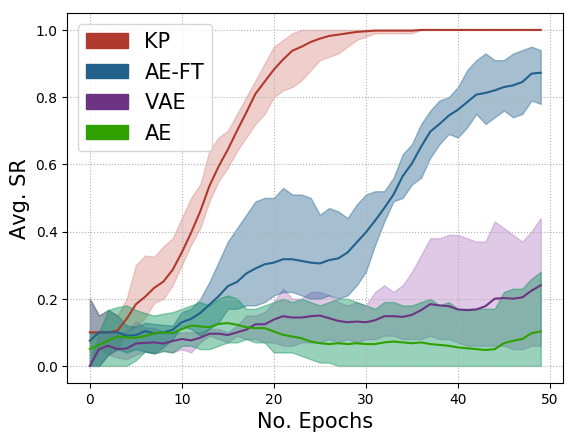}
    \caption{Success rates of \textit{UR5Grasp} env-2}
    \label{f1}
  \end{subfigure}
    \begin{subfigure}[b]{0.28\linewidth}
    \includegraphics[width=\textwidth]{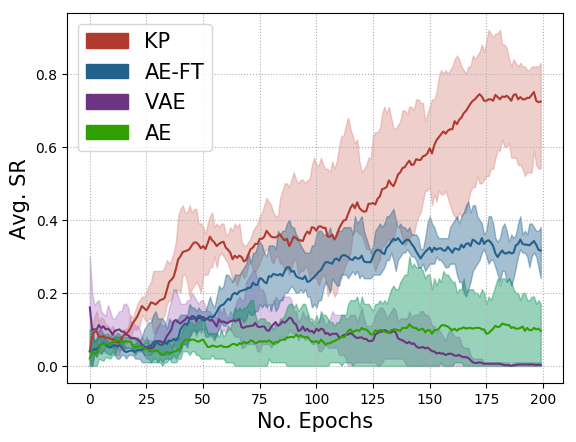}
    \caption{Success rates of \textit{UR5Push} env}
    \label{f1}
  \end{subfigure}
  
  \begin{subfigure}[b]{0.28\linewidth}
    \includegraphics[width=\textwidth]{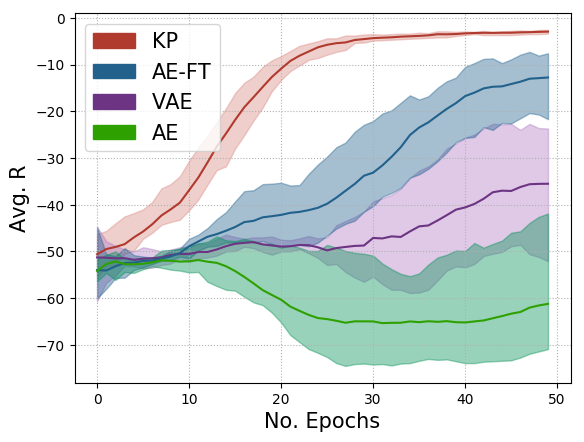}
    \caption{Average returns of \textit{UR5Grasp} env-1}
    \label{f2}
  \end{subfigure}
    \begin{subfigure}[b]{0.28\linewidth}
    \includegraphics[width=\textwidth]{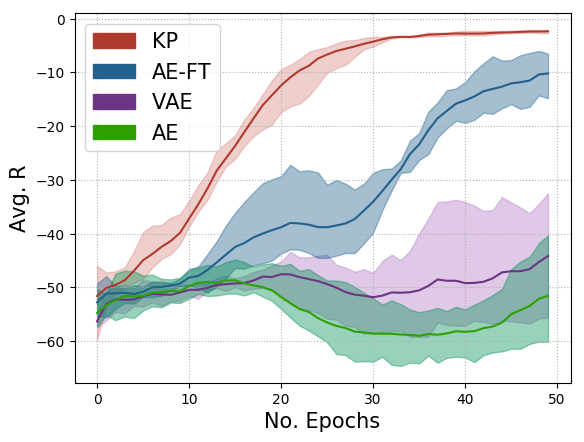}
    \caption{Average returns of \textit{UR5Grasp} env-2}
    \label{f2}
  \end{subfigure}
  \begin{subfigure}[b]{0.28\linewidth}
    \includegraphics[width=\textwidth]{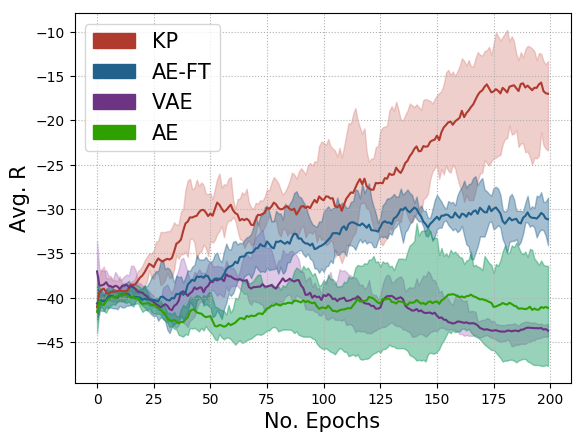}
    \caption{Average returns of \textit{UR5Push} env}
    \label{f2}
  \end{subfigure}
  \caption{Performance Benchmark of keypoint-base, VAE-based and AE-based methods for extracting features in the raw image input for the \gls{rl} module. The first row is the success rate of the manipulation task, while the second row is the corresponding episodic return. The comparisons are based on \textit{UR5Grasp} env-1, \textit{UR5Grasp} env-2, \textit{UR5Push} env. Each task has been run 10 times to plot the mean and variance.}
  \label{fig:benchmark}
  \vspace{-5.0mm}
\end{figure*}

\subsection{Setup}

We use PyBullet simulator \cite{coumans2019} for the training and simulation, with OpenAI gym \cite{openaigym16} interface. Our RL algorithm is implemented based on Stable Baselines \gls{rl} framework \cite{stable-baselines}. The environments we train and test the policy on are \textit{UR5Grasp} and \textit{UR5Push} implemented with PyBullet simulator. All simulations are run 10 times to plot the mean and variance in the figures in Section \ref{sect:res}. Actual experiments are carried out on a Universal Robot UR5 with a Robotiq 2F-85 gripper and an Intel Realsense D435 camera. The camera is placed at a fixed location at about $45 \degree$ view of the workspace; camera calibration is not required for our system. The real-world experiment setup and inference pipeline is shown in Fig. \ref{figure:exp}.

\subsection{Results}

\subsubsection{Keypoints Representation}

As visualized in Fig. \ref{figure:kp_sample}, the keypoints are constrained on either the target or the gripper, with the designed geometric loss. In addition, the keypoints extraction module also demonstrates robust tracking performance with real-world images, with the model trained only in simulated environment. Thus, with these characteristics, keypoints representation is able to reduce the input state dimensions effectively, and help to learn the control policy robustly. 


\subsubsection{Results Comparison}\label{sect:res}

We validate the proposed method on three different robotic manipulation scenarios including grasping and pushing tasks. The simulation setups of the three tasks are shown in Fig \ref{figure:tasks}. We compare our methods with the methods that use Variational Autoencoder (VAE) and Autoencoder (AE) as feature extraction modules. We use the same encoding network architecture (without expanding layer for keypoints) for VAE and AE, a pre-trained and fine-tuned ResNet18~\cite{he2016deep} for AE-FT.
All feature extraction modules are trained with the same simulation image data and fixed during the RL training stage.

As shown in Figure \ref{fig:benchmark}, the proposed \gls{rl} framework with keypoints representation significantly outperforms other methods with higher success rate and better training efficiency in all three tasks. Taking grasping task in Figure~\ref{fig:benchmark} for example, the proposed keypoints based RL methods started to converge after 20 epochs and achieves almost 100\% success rate. Moreover, keypoints-based RL method demonstrates better stability over other methods in the grasping tasks. Note that the push task is more challenging compared to grasp task, because the vision-based pushing scenario only uses the camera image to evaluate the pushing process.

\subsubsection{Number of Keypoints}
\begin{figure}[!thbp]
\centering
  \begin{subfigure}[b]{0.45\linewidth}
    \includegraphics[width=\textwidth]{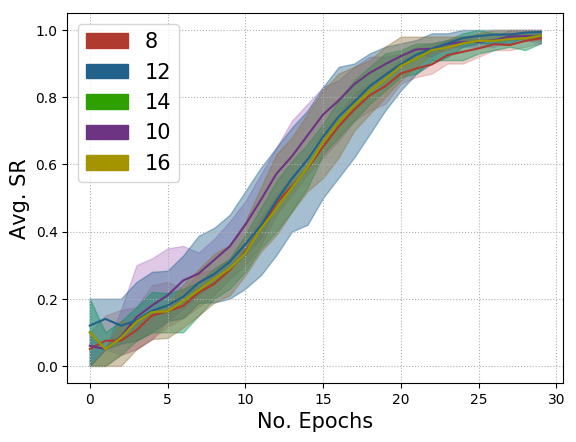}
    \caption{Success rates}
    \label{f1}
  \end{subfigure}
  \begin{subfigure}[b]{0.45\linewidth}
    \includegraphics[width=\textwidth]{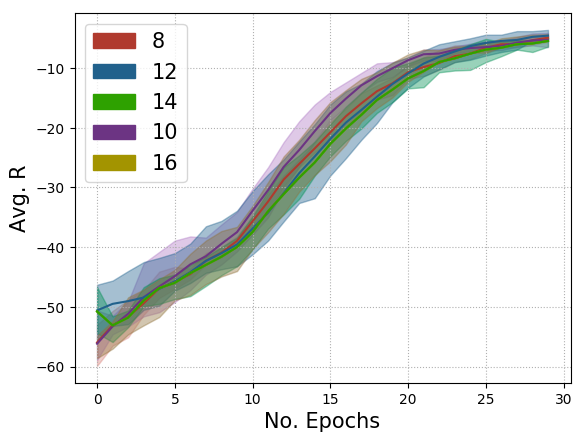}
    \caption{Average returns}
    \label{f2}
  \end{subfigure}
  \caption{Performance comparison with different numbers (${8, 10, 12, 14}$) of keypoints on grasping task}
  \label{fig:kp_compare}
  \vspace{-5.0mm}
\end{figure}

We evaluate the performance with different numbers (${8, 10, 12, 14}$) of keypoints, as shown in Fig. \ref{fig:kp_compare}. We find that the algorithm with 12 keypoints achieve best results in the grasping task. Nevertheless, the performance differences are minor with different numbers of keypoints, within a reasonable range. The results indicate that the performance of the proposed method is robust to keypoint numbers.

\subsubsection{Generalization and Transfer Learning}
\begin{figure}[!thbp]
\centering
    \includegraphics[width=0.4\textwidth]{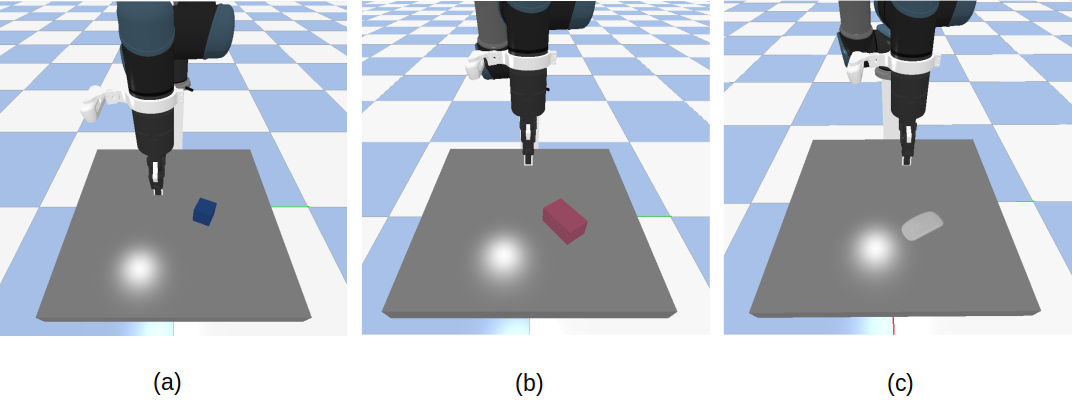}
    \caption{Generalization and transfer learning experiment setup: (a) Cube (b) Rectangle (c) Computer mouse}
    \label{figure:general_sim}
    \vspace{-3.0mm}
\end{figure}

\begin{figure}[!thbp]
\centering
  \begin{subfigure}[b]{0.45\linewidth}
    \includegraphics[width=\textwidth]{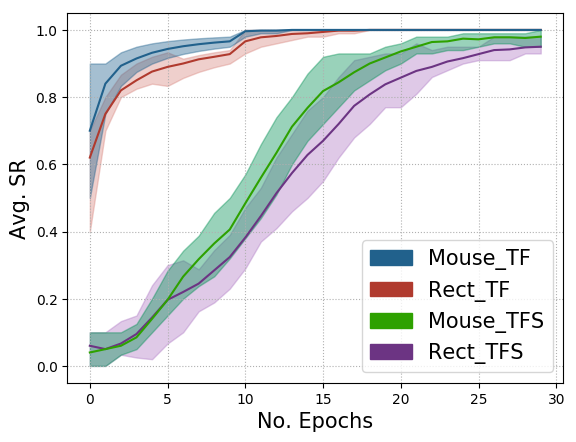}
    \caption{Success rates}
    \label{fig:generalization:f1}
  \end{subfigure}
  \begin{subfigure}[b]{0.45\linewidth}
    \includegraphics[width=\textwidth]{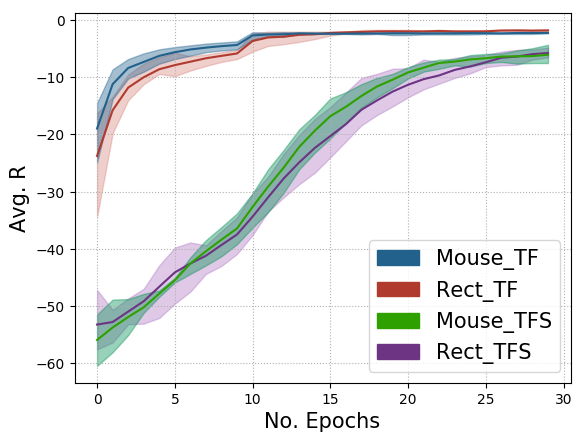}
    \caption{Average returns}
    \label{fig:generalization:f2}
  \end{subfigure}
  \caption{Generalization and transfer learning: comparison of performance on grasping rectangle on transfer learning (Rect\_TF) and training from scratch (Rect\_TFS), as well as grasping mouse on transfer learning (Mouse\_TF) and training from scratch (Mouse\_TFS)}
  \label{fig:generalization}
  \vspace{-5.0mm}
\end{figure}

As discussed in \cite{julian2020never, lee2020making, wang2019efficient}, when deploying a trained policy to new manipulation target object, the model performance usually drops. Therefore, we also investigate the task generalization capabilities of the proposed keypoints-based \gls{rl} methods. For this study, we train the policy using a cube as a primitive manipulation target, then conduct a transfer learning on a rectangle and a computer mouse, as shown in Fig. \ref{figure:general_sim}. Fig. \ref{fig:generalization} compares the results of success rates and average returns of the TFS (training from scratch) and TF (transfer learning from the pre-trained policy). The pre-trained policy shows good initial success rate and converges faster. The generalization capabilities indicate that we could train the \gls{rl} model on primitive geometries to achieve good efficiency for transferring the policy to new target object. 

\subsection{Real-world Experiments}

\begin{figure}[!thbp]
\centering
    \includegraphics[width=0.4\textwidth]{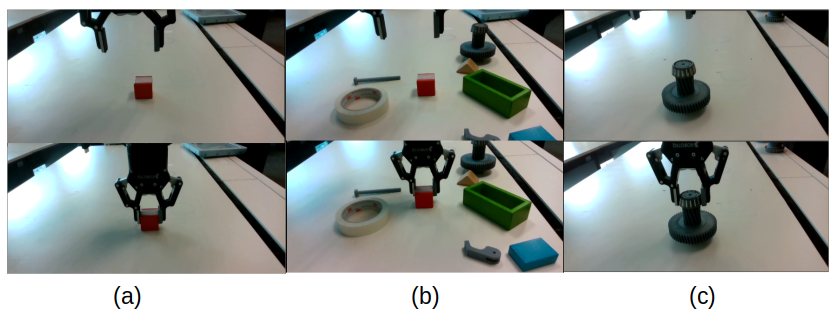}
    \caption{Experiment process from the camera viewpoint in three scenarios: (a) GB (b) GB-N (c) GT}
    \label{figure:exp_process}
    \vspace{-5.0mm}
\end{figure}

\begin{figure}[!thbp]
\centering
    \includegraphics[width=0.4\textwidth]{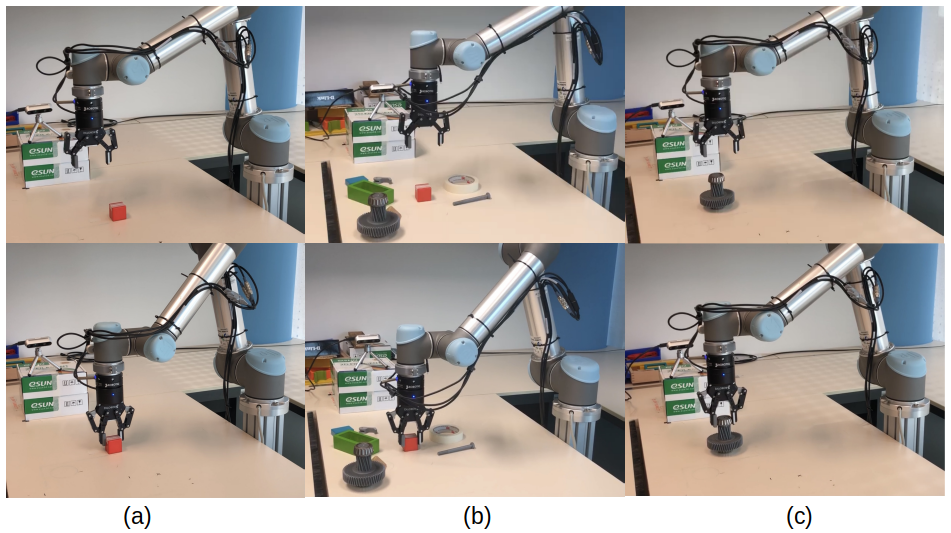}
    \caption{Experiment process from the third-person viewpoint in three scenarios: (a) GB (b) GB-N (c) GT}
    \label{figure:exp_process_cam}
    \vspace{-5.0mm}
\end{figure}

We also conduct real-world experiment directly using our policy model trained in the simulator, as shown in Fig. \ref{figure:exp}. We evaluate our methods in three scenarios, GB (grasping the block), GB-N (grasping the block in a noisy workplace) and GT (grasping the gearbox transfer component). The experiment processes are shown in Fig. \ref{figure:exp_process} and Fig. \ref{figure:exp_process_cam}.
The success rate of three scenarios are shown in Table. \ref{tab:sr_exp}. In general, the proposed keypoints-based RL method demonstrates good success rate, as well as stability and generalization capabilities in real-world experiments. However, some failure cases are still observed, which are mainly due to real-world environment factors such as noisy workplace, camera distortion and noise, as well as slippery surface.

\begin{table}[htbp]
    \caption{Success rate of real-world experiments.}
    \small
    \label{tab:sr_exp}
    \centering
    \resizebox{\columnwidth}{!}{
    \begin{tabular}{ c c c c c}
        \toprule
        \multirow{1}{*}{Scenario} & GB & GB-N & GT \\
        \midrule
        Success rate & 80\% (8/10) & 80\% (8/10) & 70\% (7/10)  \\
        \midrule

    \end{tabular}
    }
    \vspace{-5.0mm}
\end{table}

\subsection{Discussion}

The proposed \gls{rl} framework with keypoints representation demonstrates reliable manipulation behaviors in different robotic manipulation tasks, by encoding the geometric information for manipulation task efficiently. As designed in our loss function, the keypoints are forced to fall into the gripper and target object area, through the available ground truth to describe their geometric features. 
In addition, unlike the keypoints representations in \cite{gao2019kpam}, \cite{qin2019keto} that mainly encode robotic task information, our keypoints representation encode geometric information and relationship of the tool and the target in robotic manipulation, and thus it is lightweight and easy to train. In addition to the robust keypoints representation, we also include domain randomization and adversarial examples to improve the sim-to-real transfer. Moreover, we only use grey-scale images in the training and real-world scenarios to minimize the information gap and possible texture information in the real-world.

\section{Conclusion} \label{sec:conclusion}

We proposed an end-to-end reinforcement learning method with robust keypoints representation in robotic manipulation tasks. The keypoints represent compact geometric information from the camera input image, which is more explainable and robust compared to other traditional latent space representations. Moreover, using the keypoints representation trained with data augmentation, domain randomization and adversarial examples, we are able to achieve zero-shot sim-to-real transfer in real-world robotic manipulation tasks with good success rate. Based on the results in this work, we believe that using simple and compact representations such as keypoints to encode the geometric information in manipulation tasks could be promising to improve the performance of the end-to-end learning and sim-to-real transfer for robotic manipulation tasks.

\section*{Acknowledgement}

\noindent This research is supported by A*STAR, Singapore, under its AME Programmatic Funding Scheme (Project \#A18A2b0046), and Alibaba Group through Alibaba Innovative Research (AIR) Program.

\bibliography{robotrl}

\end{document}